\documentclass{article}
\usepackage{spconf,amsmath,graphicx}
\usepackage[ruled, vlined, linesnumbered]{algorithm2e}
\usepackage{setspace}
\usepackage{ragged2e} 
\usepackage{booktabs,makecell, multirow, tabularx}
\usepackage{subfigure}
\usepackage{graphicx}
\usepackage{cite}
\usepackage[T1]{fontenc}
\usepackage{stfloats}
\usepackage{bm}

\newenvironment{sequation}{\begin{equation}\small}{\end{equation}}

\title{MetaDefa: Meta-learning based on Domain Enhancement and Feature Alignment for Single Domain Generalization}

\name{Can Sun$^{*}$,Hao Zheng$^{*}$,Zhigang Hu$^{\S}$,Liu Yang,Meiguang Zheng,Bo Xu\thanks{$^{*}$Co-first authors. $^{\S}$ Zhigang Hu is the corresponding author.}}
        
\address{School of Computer Science and Engineering,
    Central South University,
    China}

\begin{document}

\maketitle

\begin{abstract}

    The single domain generalization(SDG) based on meta-learning has emerged as an effective technique for solving the domain-shift problem. However, the inadequate match of data distribution between source and augmented domains and difficult separation of domain-invariant features from domain-related features make SDG model hard to achieve great generalization. Therefore, a novel meta-learning method based on domain enhancement and feature alignment (MetaDefa) is proposed to improve the model generalization performance. First, the background substitution and visual corruptions techniques are used to generate diverse and effective augmented domains. Then, the multi-channel feature alignment module based on class activation maps and class agnostic activation maps is designed to effectively extract adequate transferability knowledge. In this module, domain-invariant features can be fully explored by focusing on similar target regions between source and augmented domains feature space and suppressing the feature representation of non-similar target regions. Extensive experiments on two publicly available datasets show that MetaDefa has significant generalization performance advantages in unknown multiple target domains.

\end{abstract}

\begin{keywords}
    Single domain generalization, Domain enhancement, Feature alignment, Meta-learning
\end{keywords}

\section{Introduction}
Deep neural networks driven by numerous labeled samples has made remarkable progress in a wide range of computer vision tasks \cite{mi2023multiple}. However, due to the significant data distribution differences between the source and target domains, the model performance will decrease apparently\cite{wang2018deep}, which is known as the domain-shift problem.

Single domain generalization can effectively solve the domain-shift problem. The SDG method improves the model generalization performance by training the model in single source domain to learn transferability knowledge and applying the knowledge learned to the unknown multiple target domains \cite{wang2022generalizing,xu2023simde}. Some scholars researched SDG based on the feature representation to reduce the feature space difference between source and augmented domains, making the model more focused on domain-invariant features \cite{shu2021open,segu2023batch,hou2023learning}. However, the limited performance of the feature representation method itself brings unsatisfactory feature alignment and loss of source domain information. On the other hand, in order to adapting quickly to new tasks and classes in unknown target domains, many researchers started to apply meta-learning to SDG by dividing the source domain into several virtual train domains and virtual test domains. For example, some methods \cite{qiao2020learning,balaji2018metareg,li2019feature,zhao2021learning} make efforts to make the model learn sufficient prior knowledge and shared experience by simulating the generalization step in the training process. However, these methods are also difficult to generate more adaptable augmented domains and extract adequate transferability knowledge. Even worse, the model suffers from the problem of neglecting target category feature representation and insufficient suppression of non-target categories.

Hence, to solve the above deficiencies, a meta-learning method based on domain enhancement and feature alignment for single domain generalization is designed. First, a domain enhancement module using background substitution and visual corruptions techniques is proposed, which considers the reality of the enhanced domain while introducing more variation and uncertainty. Second, the multi-channel feature alignment module is designed to reduce the gap between the target category regions in both the source and augmented domains feature space and squeeze non-target category areas.

The main contributions of this paper are listed as follows.

a) The domain enhancement module based on background substitution and visual corruptions is proposed to generate diverse and effective styles of augmented domains to adequately simulate unknown domain distributions

b) The multi-channel feature alignment module suppresses image CAAM approaching CAM and enlarges image secondary regions respectively in each iteration of meta-training and meta-testing. This achieves intra-class compactness and inter-class separability by imposing consistency on the CAM of the original and enhanced images.


c) Extensive experiments are conducted on two benchmark datasets, and the experimental results demonstrated the superior comprehensive performance of MetaDefa.

\begin{figure*}[htb]
    \setlength{\abovecaptionskip}{0.cm}
    \setlength{\belowcaptionskip}{0.cm}
    \centerline{\includegraphics[scale=0.36]{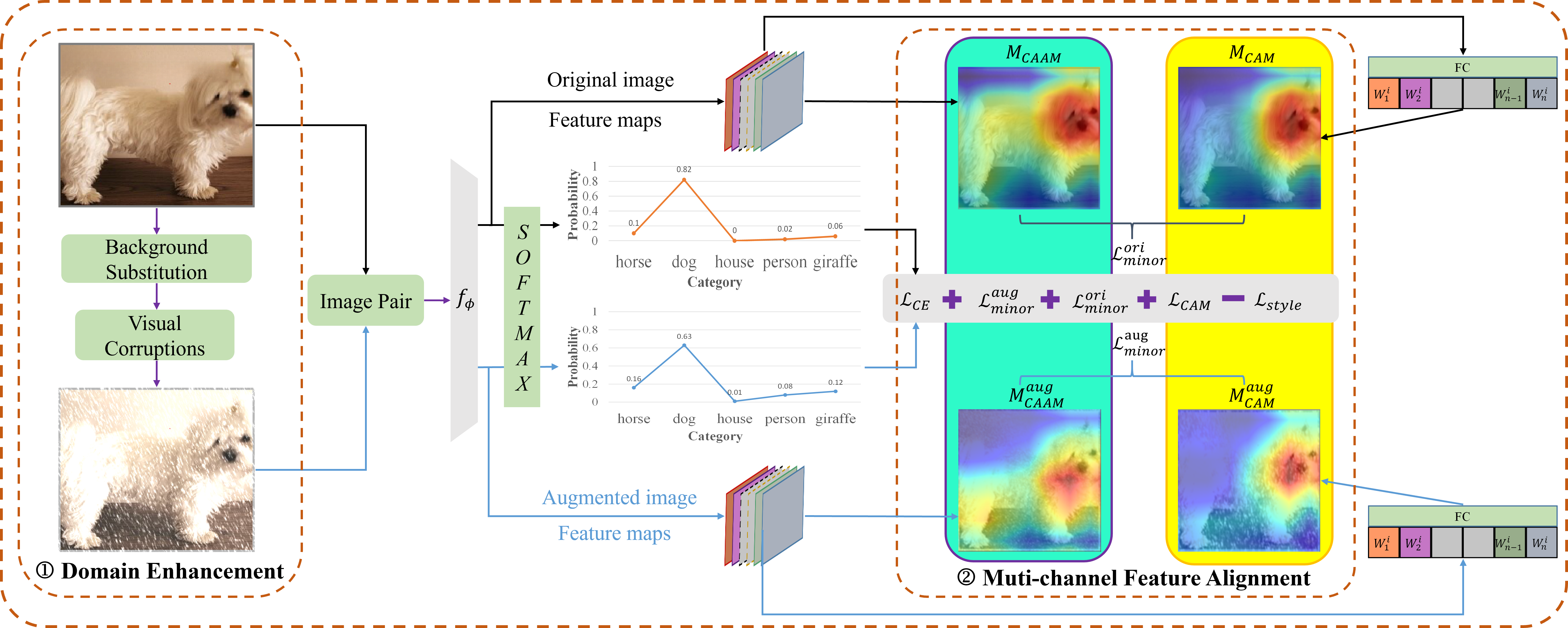}}
    \caption{The domain enhancement and multi-channel feature alignment module of MetaDefa. The function $f_{\phi}$ denotes the model with parameter $\phi$. The $w_{k}^{i}$ represents the weight of the kth feature map in the fully connected layer with respect to category $i$.} \label{fig2}
\end{figure*}

\section{Method}

In the context of meta-learning for SDG, consider a single source domain \emph{S} and multiple target domains \emph{T}. The domain \emph{S} will be partitioned into virtual train domain $\emph{S}_{train}$ and virtual test domain $\emph{S}_{test}$. The function $f_{\phi}: S_{x} \rightarrow \mathcal{Y}$ is introduced to map the input image \emph{X} from \emph{S} to a label-like heat vector \emph{Y}, and the parameter $\phi$ need to be learned. During each iteration, the meta-train stage is entered first. The model starts training on $\emph{S}_{train}$, where the loss and gradients are calculated, and the parameter $\phi$ are updated as $\phi \longrightarrow \hat{\theta}^{i}$. Subsequently, the meta-test stage is executed. Here, the model is trained on $\emph{S}_{test}$ utilizing the updated parameter $\hat{\theta}^{i}$ to calculate $\mathcal{L}\left(f_{\widehat{\theta}^{i}}\right)$. The gradients are then computed and saved. Finally, this entire process is repeated 'n' times, and all the stored gradients are accumulated and used to update the initial parameter $\phi$.

Fig. 1 shows the domain enhancement and multi-channel feature alignment module of MetaDefa. The algorithmic details of MetaDefa are shown in Algorithm 1.

\subsection{Domain Enhancement}
To adequately match the data distribution between $\emph{S}_{train}$ and virtual augmented train domain $S_{train}^{aug}$, $\emph{S}_{test}$ and virtual augmented test domain $S_{test}^{aug}$, taking diversity and effectiveness into consideration, the new domain enhancement module generates optimal enhanced domains through background substitution and visual corruptions techniques.

$\textbf{a) Background substitution:}$ Prior research has demonstrated that data augmentation methods that prioritize diversity over effectiveness will lead to a decline in performance \cite{deng2021labels}. Background substitution techniques are performed to ensure effectiveness. During the replacement process, we initially utilize instance mask annotations to identify the object region within one image and keep it unchanged. Another image from $\emph{S}_{train}$ of a different class is selected to extract a random patch. Finally, the image background is replaceed with the random patch to produce a valid enhanced image.

$\textbf{b) Visual corruptions:}$
The diversity of images brought about by visual corruptions dramatically boost the model's generalization capabilities when dealing with unknown multiple target domain. To expand the difference in data distribution between $\emph{S}_{train}$ and $S_{train}^{aug}$, $\emph{S}_{test}$ and $S_{test}^{aug}$, a minimum threshold is designed for fundamental visual corruptions. Only visual corruptions with a randomly impairment probability above the threshold can be executed.

\subsection{Multi-channel Feature Alignment}
The transferable knowledge between $\emph{S}_{train}$ and $S_{train}^{aug}$, $\emph{S}_{test}$ and $S_{test}^{aug}$ observably influences model generalization. The multi-channel feature alignment module including focusing on domain-invariant features and inhibiting domain-related features is used to extract adequate transferability knowledge.

$\textbf{a) Focus on domain-invariant features:}$ The class activation maps(CAM) is known for visualizing the spatial regions of feature maps \cite{zhou2016learning}. Different from the CAM-loss \cite{wang2021towards}, this paper aims to constrain the model to find consistent and generic visual cues within various views of the same input image by minimizing the distance between CAMs of $\emph{S}_{train}$ and $S_{train}^{aug}$. The model can then reuse these cues when dealing with unfamiliar target domains.

\begin{algorithm}[htb]
    \setstretch{0.5}
    \SetAlgoLined
    \caption{MetaDefa}\label{algorithm}
    \KwIn{source domain $S$, learning rate $lr$, model parameter $\phi$, hyperparameters $\beta, \lambda_{1}, \lambda_{2}$.}
    \KwOut{parameter of the finished training model $\phi$}
    Extract data from source domain $S$ for building task pool with $N$ size, which contains $S_{train}, S_{test}$. \\
    Randomly initialize $\phi$ \\

    \For{epoch to epochs}{
    Sample $n$ tasks $T_{i}$ from task pool \\

    \For{all $T_{i}$}{

    \tcp{Meta-train stage}

    Carry out Background Substitution on $S_{train}$ \\
    \% Maintain Effectiveness \\
    Carry out Visual Corruptions on $S_{train}$ \\
    \% Maintain Diversity \\
    Output $S_{train}$ and $S_{train}^{aug}$ and compute the loss: $\mathcal{L}_{T_{i}}\left(f_{\widehat{\theta}^{i}}\right)$ \\
    Compute the adapted parameters with gradient descent: $\hat{\theta}^{i}=\phi - lr \cdot \nabla_{\phi} \mathcal{L}_{T_{i}}\left(f_{\widehat{\theta}^{i}}\right)$ \\

    \tcp{ Meta-test stage}

    Carry out Background Substitution on $S_{test}$ \\
    Carry out Visual Corruptions on $S_{test}$ \\
    Output $S_{test}$ and $S_{test}^{aug}$ and compute the loss: $\mathcal{L}_{T_{i}}\left(f_{\widehat{\theta}^{i}}\right)$  \\
    Compute the gradient: $ \nabla_{\widehat{\theta}^{i}} \mathcal{L}_{T_{i}}\left(f_{\widehat{\theta}^{i}}\right) $  \\

    }
    \tcp{ Actual parameter updates}
    Update $\phi=\phi-\beta \cdot \nabla_{\phi} \sum_{1}^{n} \mathcal{L}_{T_{i}}\left(f_{\widehat{\theta}^{i}}\right) / n$

    }
\end{algorithm}

For a given image, let $f_{k}(x, y)$ represent the activation value of the kth feature map at the spatial location $(x,y)$. A global average pooling operation is executed: $F_{k}=\frac{1}{H \times W} \sum_{x, y} f_{k}(x, y)$. For a given class $i$, a softmax operation is applied to obtain $z_{i}$ as: $z_{i}=\sum_{k} w_{k}^{i} F_{k}$. Combining the expressions for $F_{k}$ and $z_{i}$ yields: $z_{i} = \frac{1}{H \times W} \sum_{x, y} \sum_{k} w_{k}^{i} f_{k}(x, y)$.  Define $CAM_{i}$ as the class activation maps of class $i$, where each spatial element is represented as $\operatorname{CAM}_{i}(x, y)=\sum_{k} w_{k}^{i} f_{k}(x, y)$.

To fully explore domain-invariant features, the $\mathcal{L}_{CAM}$ loss is formulated using the Jensen-Shannon divergence:

\begin{sequation}
    \setlength{\abovedisplayskip}{0.1cm}
    \setlength{\belowdisplayskip}{0.1cm}
    \begin{aligned}
        \mathcal{L}_{C A M}\left(M_{C A M}, M_{C A M}^{aug}, i\right)=D_{J S}\left(M_{C A M} \| M_{C A M}^{aug}\right)
    \end{aligned}
\end{sequation}

where $M_{C A M}$ represents the CAM of $\emph{S}_{train}$ for a given class $i$, $M_{C A M}^{aug}$ represents the CAM of $S_{train}^{aug}$.

$\textbf{b) Inhibit domain-related features:}$ As observed in Fig.2(b)(e), Since the CAMs of target and non-target categories may overlap in some areas, the model that only consider CAMs has limitations in capturing the intricate relationships between target and non-target categories. The class agnostic activation maps (CAAM) presents more significant activation regions and richer features than CAM, as shown in Fig. 2(c)(f), where each spatial element is expressed as: $\operatorname{CAAM}(x, y)=\sum_{k} f_{k}(x, y)$. Encouraging CAAM to closely align with CAM of the target category to suppress the expression of non-target category features. $\mathcal{L}_{minor}^{ori}$ and $\mathcal{L}_{minor}^{aug}$ are formulated as:

\begin{sequation}
    \setlength{\abovedisplayskip}{0.1cm}
    \setlength{\belowdisplayskip}{0.1cm}
    \begin{aligned}
        \mathcal{L}_{\text {minor }}^{\text {ori }}=\frac{1}{H \times W} \sum_{x, y}\left\|M_{C A A M}-M_{C A M}\right\|_{l_{1}}
    \end{aligned}
\end{sequation}

\begin{sequation}
    \setlength{\abovedisplayskip}{0.1cm}
    \setlength{\belowdisplayskip}{0.1cm}
    \begin{aligned}
        \mathcal{L}_{\text {minor }}^{ \text { aug }}=\frac{1}{H \times W} \sum_{x, y}\left\|M_{C A A M}^{aug}-M_{C A M}^{aug}\right\|_{l_{1}}
    \end{aligned}
\end{sequation}

$M_{C A A M}$ and $M_{C A A M}^{aug}$ denote the CAAM of $S_{train}$ and $S_{train}^{aug}$, respectively. Compared with the CAM-loss \cite{wang2021towards}, in order to further enhance the model's perception and suppression of non-target categories, the secondary areas between $S_{train}$ and $S_{train}^{aug}$ should be as different as possible. The $\mathcal{L}_{style}$ loss is formulated as follows:

\begin{figure}[tb]
    \flushleft
    \subfigure[]{
        \includegraphics[width=1.2cm]{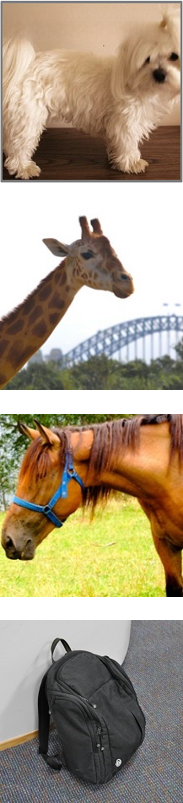}}
    \subfigure[]{
        \includegraphics[width=1.2cm]{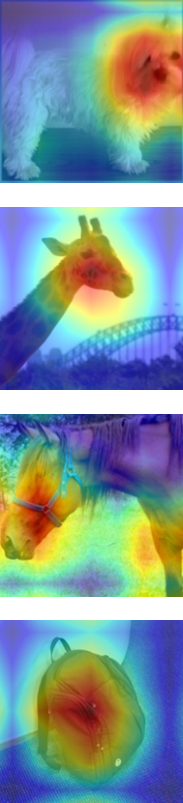}}
    \subfigure[]{
        \includegraphics[width=1.2cm]{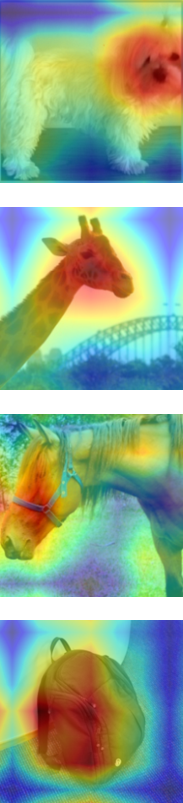}}
    \subfigure[]{
        \includegraphics[width=1.2cm]{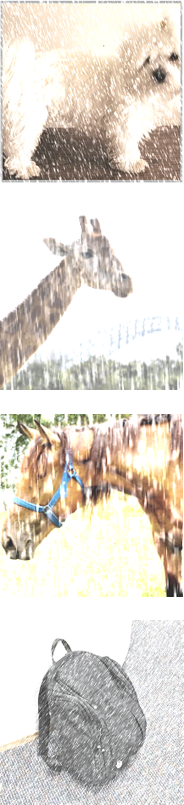}}
    \subfigure[]{
        \includegraphics[width=1.2cm]{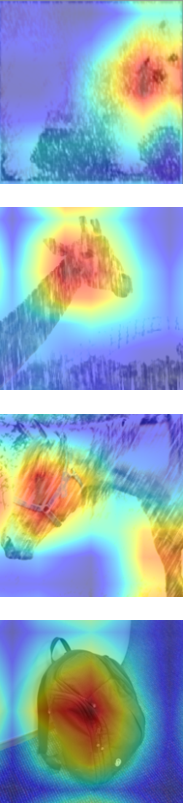}}
    \subfigure[]{
        \includegraphics[width=1.2cm]{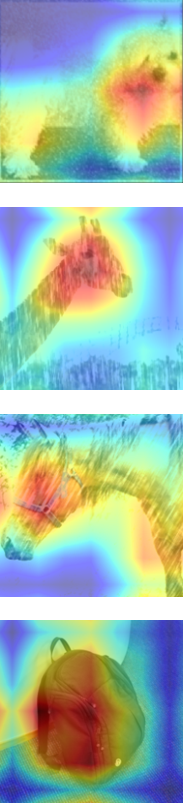}}
    \caption{Examples illustrate the difference between CAM and CAAM. (a) represents the $S_{train}$, (b) and (c) represent the $M_{C A M}$ and $M_{C A A M}$. Similarly, (d) represents the $S_{train}^{aug}$, (e) and (f) represent the $M_{C A M}^{aug}$ and  $M_{C A A M}$.}
    \label{Fig:weight}
\end{figure}

\begin{sequation}
    \setlength{\abovedisplayskip}{0.1cm}
    \setlength{\belowdisplayskip}{0.1cm}
    \begin{aligned}
        \mathcal{L}_{\text {style }}\left(\mathcal{L}_{\text {minor }}^{\text {ori }}, \mathcal{L}_{\text {minor }}^{\text {aug }}, i\right)=\left|\mathcal{L}_{\text {minor }}^{\text {ori }}-\mathcal{L}_{\text {minor }}^{\text {aug }}\right|
    \end{aligned}
\end{sequation}

According to (1)(2)(3)(4), by introducing the hyperparameters $\lambda_{1}$ and $\lambda_{2}$, the final objective function is defined as:

\begin{sequation}
    \setlength{\abovedisplayskip}{0.1cm}
    \setlength{\belowdisplayskip}{0.1cm}
    \begin{aligned}
        \mathcal{L}=\mathcal{L}_{C E}+\lambda_{1}\left(\mathcal{L}_{C A M}+\mathcal{L}_{\text {minor }}^{\text {ori }}+\mathcal{L}_{\text {minor }}^{\text {aug }}\right)-\lambda_{2} \mathcal{L}_{\text {style }}
    \end{aligned}
\end{sequation}

\section{Experiments}

\subsection{Datasets and Setup}
We conduct an extensive evaluation of MetaDefa using two benchmark datasets. The Office-Caltech-10 dataset \cite{gong2012geodesic} comprises 2533 samples distributed across four domains. The Office31 \cite{saenko2010adapting} includes 4110 images divided into 31 categories, collected from three domains. The Office31 is more challenging due to its larger number of categories compared to Office-Caltech-10 with only 10 categories.

For all settings, RGB images is uniformly resized to 224$\times$224. The domain DSLR is chosen as the single source domain. We adopt ResNet-18 pre-trained on ImageNet. The learning rate, batch size and training epoch are set to 4$\times10^{-3}$, 128 and 30, respectively. To this end, we repeat all experiments five times and take their average as the final results.

\subsection{Baseline and Comparison Methods}
To assess the effectiveness of MetaDefa, comparisons are made in terms of both domain enhancement and feature alignment. These comparisons includes (1) baseline: meta-learning using only cross-entropy loss without any domain enhancement, (2) CutOut \cite{devries2017improved} and RandAugment(RandAug) \cite{cubuk2020randaugment} using advanced domain enhancement techniques, while other module settings are consistent with MetaDefa; (3) L2D \cite{wang2021learning} based on semantic consistency to align domain-invariant features, (4) ACVC \cite{cugu2022attention} based on class activation map but not suppress non-target categories,  and (5) CAM-loss \cite{wang2021towards}: Constraining CAAM to lean on CAM.

\begin{table}[tb]
    \centering
    \caption{Comparison of MetaDefa(Ours) on the office-Caltech-10 benchmark with other methods.}
    \label{table1}
    \scalebox{0.7}{
        \begin{tabular}{c|c|c|c|c}
            \hline
            \hline

            Method                  & Amazon              & Caltech             & Webcam              & AVG                 \\

            \hline
            Baseline                & 84.05±3.63          & 77.24±2.57          & 97.03±1.23          & 86.11±1.73          \\

            \hline
            CutOut                  & 84.93±2.09          & 77.63±2.12          & 96.20±0.90          & 86.25±1.46          \\

            \hline
            RandAug                 & 83.34±2.20          & 77.26±1.20          & 97.63±1.12          & 86.07±0.99          \\

            \hline
            L2D                     & 83.23±0.31          & 79.16±0.48          & 96.61±0.29          & 86.33±0.36          \\

            \hline
            ACVC                    & 86.37±2.19          & 77.72±2.40          & 97.23±1.17          & 87.21±0.93          \\

            \hline
            CAM-loss                & 86.18±0.80          & 78.59±0.68          & 97.56±0.58          & 87.44±0.26          \\

            \hline
            \textbf{MetaDefa(Ours)} & \textbf{88.44±1.69} & \textbf{80.21±0.61} & \textbf{97.97±0.42} & \textbf{88.87±0.45} \\

            \hline
            \hline
        \end{tabular}
    }
\end{table}

\begin{table}[tb]
    \centering
    \caption{Comparison of MetaDefa(Ours) on the office31 benchmark with other methods.}
    \label{table2}
    \scalebox{0.9}{
        \begin{tabular}{c|c|c|c}
            \hline
            \hline

            Method                  & Amazon              & Webcam              & AVG                 \\

            \hline
            Baseline                & 47.86±2.02          & 90.62±1.10          & 69.24±1.04          \\

            \hline
            CutOut                  & 46.89±1.77          & 90.11±0.59          & 68.50±1.11          \\

            \hline
            RandAug                 & 48.13±1.53          & 91.22±1.40          & 69.67±1.25          \\

            \hline
            L2D                     & 51.12±0.47          & 92.68±0.48          & 71.89±0.24          \\

            \hline
            ACVC                    & 50.60±2.40          & 92.10±0.69          & 71.35±1.01          \\

            \hline
            CAM-loss                & 51.15±1.68          & 92.45±0.41          & 71.80±1.04          \\

            \hline
            \textbf{MetaDefa(Ours)} & \textbf{53.19±1.01} & \textbf{92.94±1.39} & \textbf{73.06±0.82} \\

            \hline
            \hline
        \end{tabular}
    }
\end{table}

\subsection{Comparison on office-Caltech-10 and office31}
As shown in Table 1, MetaDefa consistently achieves the optimal generalization effect across the three target domains, which the accuracies are 88.44\%, 80.21\% and 97.97\%, respectively. By considering the effectiveness of enhancement, MetaDefa increases the performance by 2.62\% and 2.8\% compared to CutOut and RandAug that only care about diversity. This verifies the effectiveness of the proposed domain enhancement module. Moreover, due to MetaDefa suppresses non-target category features during feature alignment, the model gains more transferable knowledge, which results in MetaDefa improves model performance by 2.54\%, 1.66\% and 1.43\% compared to L2D, ACVC and CAM-loss.

Regarding Table 2, MetaDefa obtains the superior generalization performance on the office31 dataset.
Only using simple domain enhancement methods such as CutOut and Randug, the model's generalization performance improvement is limited, which emphasizes the importance of considering the effectiveness of enhancement. MetaDefa performs better in the face of datasets with more categories and significant domain distribution differences by inhibiting domain-related features. For instance, MetaDefa enhances model accuracy by 3.82\% on office31 and only achieves a performance improvement of 2.76\% on office-Caltech-10. Combined with Fig. 3(a)(b), MetaDefa has the competitive universality and advancement on two datasets.

\begin{table}[tb]
    \centering
    \caption{Ablation experiments of different loss items on office-Caltech-10 and office31 datasets.}
    \label{table3}
    \scalebox{0.65}{
        \begin{tabular}{c|c|c|c|c|c}
            \hline
            \hline

             & $\mathcal{L}_{CE}$ & $\mathcal{L}_{CAM}$ & $\mathcal{L}_{minor }^ {ori}+\mathcal{L}_{ minor }^{ aug }$ & $L_{style} $ & AVG                 \\

            \hline
            \multirow{4}{*}{{Office-Caltech-10}}
             & $\surd$            &                     &                                                             &              & 86.11±1.73          \\
             & $\surd$            & $\surd$             &                                                             &              & 86.60±1.39          \\
             & $\surd$            & $\surd$             & $\surd$                                                     &              & 87.01±1.61          \\
             & $\surd$            & $\surd$             & $\surd$                                                     & $\surd$      & \textbf{88.87±0.45} \\

            \hline
            \multirow{4}{*}{{office31}}
             & $\surd$            &                     &                                                             &              & 69.24±1.04          \\
             & $\surd$            & $\surd$             &                                                             &              & 71.21±0.59          \\
             & $\surd$            & $\surd$             & $\surd$                                                     &              & 72.15±0.48          \\
             & $\surd$            & $\surd$             & $\surd$                                                     & $\surd$      & \textbf{73.06±0.82} \\

            \hline
            \hline
        \end{tabular}
    }
\end{table}

\begin{figure}[tb]
    \flushleft
    \subfigure[office-Caltech-10]{
        \includegraphics[width=4.0cm]{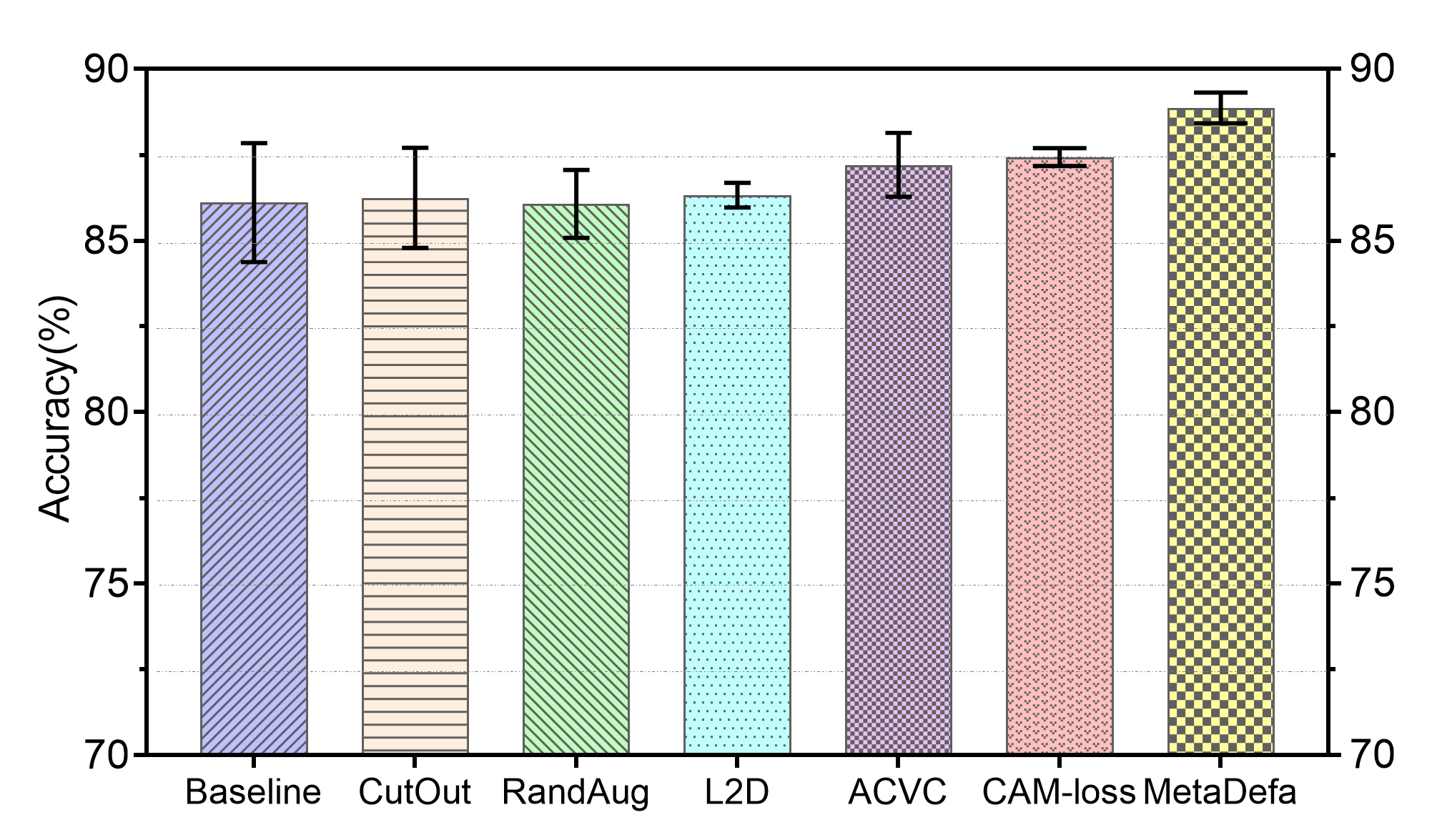 }}
    \subfigure[office31]{
        \includegraphics[width=4.25cm]{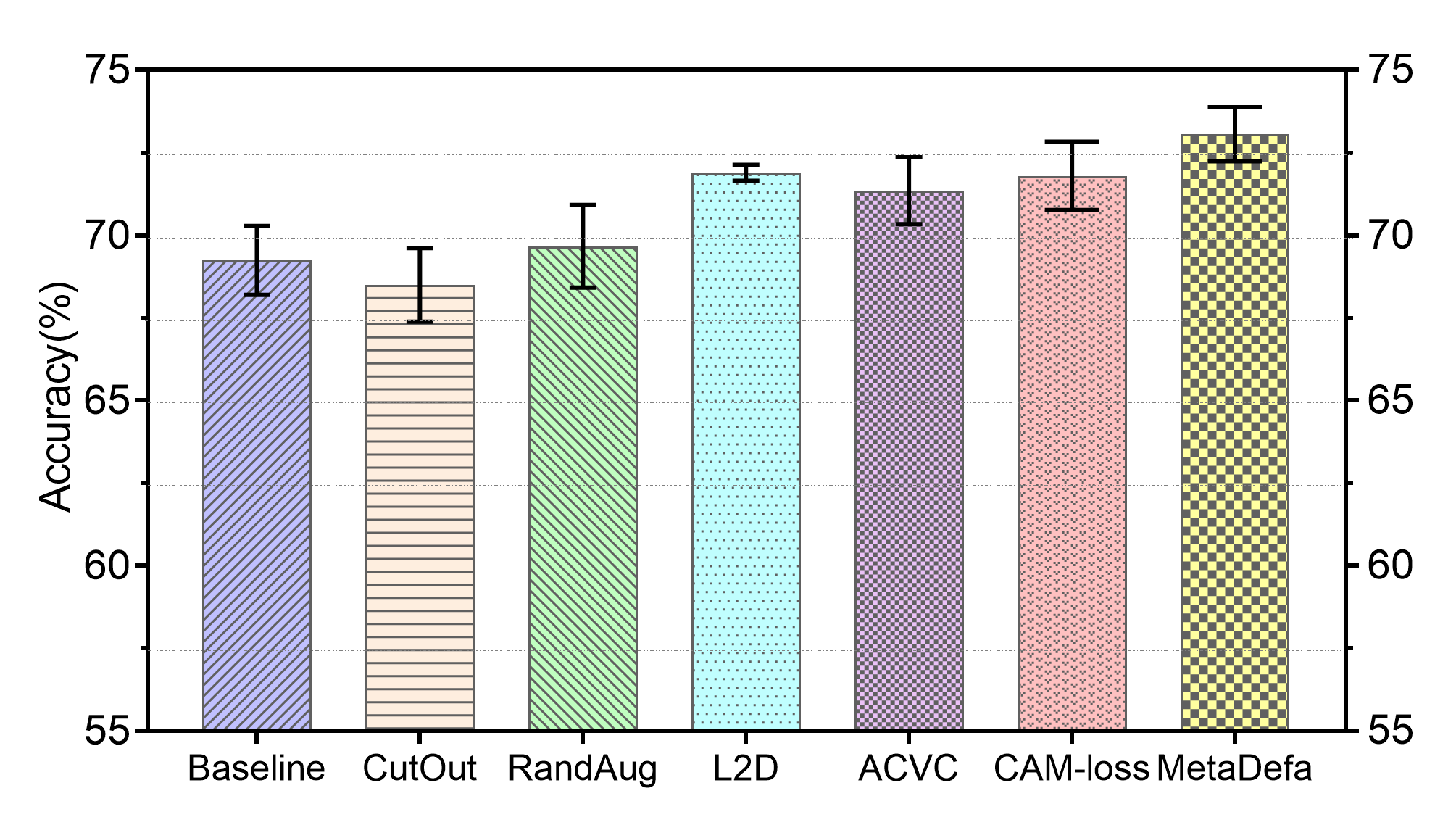 }}

    \caption{Accuracy and standard deviation of MetaDefa with advanced methods on office-Caltech-10 and office31 dataset.}
\end{figure}

\subsection{Analysis}
In MetaDefa, the different loss terms constructed based on multi-channel feature alignment are crucial. Sufficient experiments on two benchmark datasets are conducted to evaluate their effectiveness. As shown in Table 3, all the designed loss terms can improve the model prediction accuracy over the baseline. In office-Caltech-10 with a small number of categories, suppressing non-target category features will have the better performance improvement, increasing by 1.86\%. Encouraging the feature expression of the target category improves the model performance by 1.97\% in office31.

\section{Conclusion}

This paper proposes a meta-learning scheme based on domain enhancement and feature alignment. The MetaDefa focus on creating diverse and influential enhancement domains and effectively extract adequate transferability knowledge. Experimental results show that MetaDefa achieves excellent generalization performance on the benchmark datasets. In the future,
we will take into account both the model's output and feature maps to enhance the model's generalization.


\newpage

\bibliographystyle{IEEEbib}
\bibliography{strings}

\end{document}